\documentclass{article}

 \usepackage[preprint]{neurips_2026}

\usepackage{chemformula}

\usepackage[dvipsnames]{xcolor}
\newcommand{\Z}[1]{\textcolor{ForestGreen}{\text{#1}}}
\newcommand{\Qp}{\textcolor{blue}{+}}
\newcommand{\Qn}{\textcolor{red}{-}}
\newcommand{\ii}{\textsc{Inchified Invariants}}

\usepackage{soul}

\usepackage{microtype}
\usepackage{graphicx}
\usepackage{subcaption}
\usepackage{booktabs} 

\usepackage{svg}
\usepackage{graphicx}
\graphicspath{ {./fig/} }

\usepackage{array}
\usepackage{tabularx}
\usepackage{wrapfig}

\usepackage[utf8]{inputenc} 
\usepackage[T1]{fontenc}    
\usepackage{hyperref}       
\usepackage{url}            
\usepackage{booktabs}       
\usepackage{amsfonts}       
\usepackage{nicefrac}       
\usepackage{microtype}      

\hypersetup{
  colorlinks   = true, 
  urlcolor     = blue, 
  linkcolor    = blue, 
  citecolor    = blue  
}

\usepackage{amsmath}
\usepackage{amssymb}
\usepackage{mathtools}
\usepackage{amsthm}

\usepackage[capitalize,noabbrev]{cleveref}

\usepackage{algorithm}
\usepackage[noend]{algpseudocode}

\title{Aligning Molecular Graph Explanations with Chemical Identity via InChIfied Invariants}
    
\author{%
  Emanuele Guidotti \\
  University of Lugano, Switzerland\\
  \texttt{emanuele.guidotti@usi.ch} \\
  \And
  Sara Puglioli \\
  Philochem AG, Switzerland\\
  \texttt{sara.puglioli@philochem.ch} \\
}

\begin{document}

\maketitle

\begin{abstract}
Obtaining consistent explanations for machine learning on molecular graphs requires predictions and attributions to be aligned with chemical identity. However, chemically equivalent drawings of the same molecule can induce different molecular representations, leading to inconsistent predictions and explanations. Here, we introduce \ii, a class of node, edge, and graph features based on the International Chemical Identifier (InChI) and designed to be invariant under transformations that preserve chemical identity. Using one million molecular graphs from PubChem Substances, we show that \ii~produce identical representations for chemically equivalent graphs in 99.62\% of cases, whereas standard Daylight invariants do so in only 0.35\% of cases. Across MoleculeNet tasks, \ii~preserve predictive performance while significantly improving prediction consistency across alternative graph depictions of the same molecules. We further perform a quantitative attribution analysis and show that explanations produced with standard molecular featurization methods vary substantially across chemically equivalent graphs, while \ii~enforce consistent attributions by construction. We release open-source software implementing \ii, which can be used as a drop-in replacement for standard molecular graph features.\footnote{Code will be made publicly available upon publication.}
\end{abstract}

\section{Introduction}
\label{sec:intro}

Molecular graphs are a central representation in machine learning for chemistry, where atoms are modeled as nodes and bonds as edges. As predictive power increases, there is a growing emphasis on making these models explainable \cite{jimenez2020drug,vamathevan2019applications}. Experimental chemists are often skeptical of black-box predictions \cite{wellawatte2025human} and can prioritize explainability over accuracy, demanding explanations that are aligned with chemical intuition and domain knowledge \cite{chen2022machine}. Consequently, obtaining reliable explanations is now seen as a key requirement for machine-learning models in chemistry \cite{jimenez2020drug,vamathevan2019applications}.

Explanation systems should satisfy two broad criteria: explanations should (i) make sense to human users and (ii) accurately reflect the behavior of the predictive model~\cite{dasgupta2022framework}. Since the first criterion is difficult to formalize and evaluate, much of the existing literature has focused on the second, with metrics such as faithfulness becoming standard. However, whether explanations make sense to domain experts remains comparatively underexplored.

This paper addresses this first criterion in molecular machine learning. We argue that a minimal requirement for an explanation to make sense to a chemist is consistency across chemically equivalent graph representations: two drawings of the same molecule should not yield different chemical explanations. We therefore study explanation consistency across chemically equivalent graphs as a prerequisite for chemical meaningfulness. This property is necessary but not sufficient, since a meaningful explanation must be consistent, but a consistent explanation may not be chemically meaningful. Nevertheless, in the absence of widely accepted ground-truth benchmarks for chemical meaningfulness at scale, enforcing consistency with chemical identity provides a necessary first step towards more reliable explanations in molecular machine learning.

We introduce \ii: node, edge, and graph features designed to be invariant under transformations of a molecular graph that preserve chemical identity. Our construction leverages the International Chemical Identifier (InChI), a chemical structure identifier developed by the International Union of Pure and Applied Chemistry (IUPAC) to provide a standard representation of molecular identity~\cite{heller2015inchi}. InChI is designed so that alternative depictions of the same molecule map to the same identifier. We construct \ii~so that graphs with the same InChI produce the same features, while graphs with different InChIs generally produce different features. Consequently, models built on \ii~can produce predictions and attributions that are invariant across chemically equivalent graph representations.

\begin{figure}
    \centering
    \includegraphics[width=\linewidth]{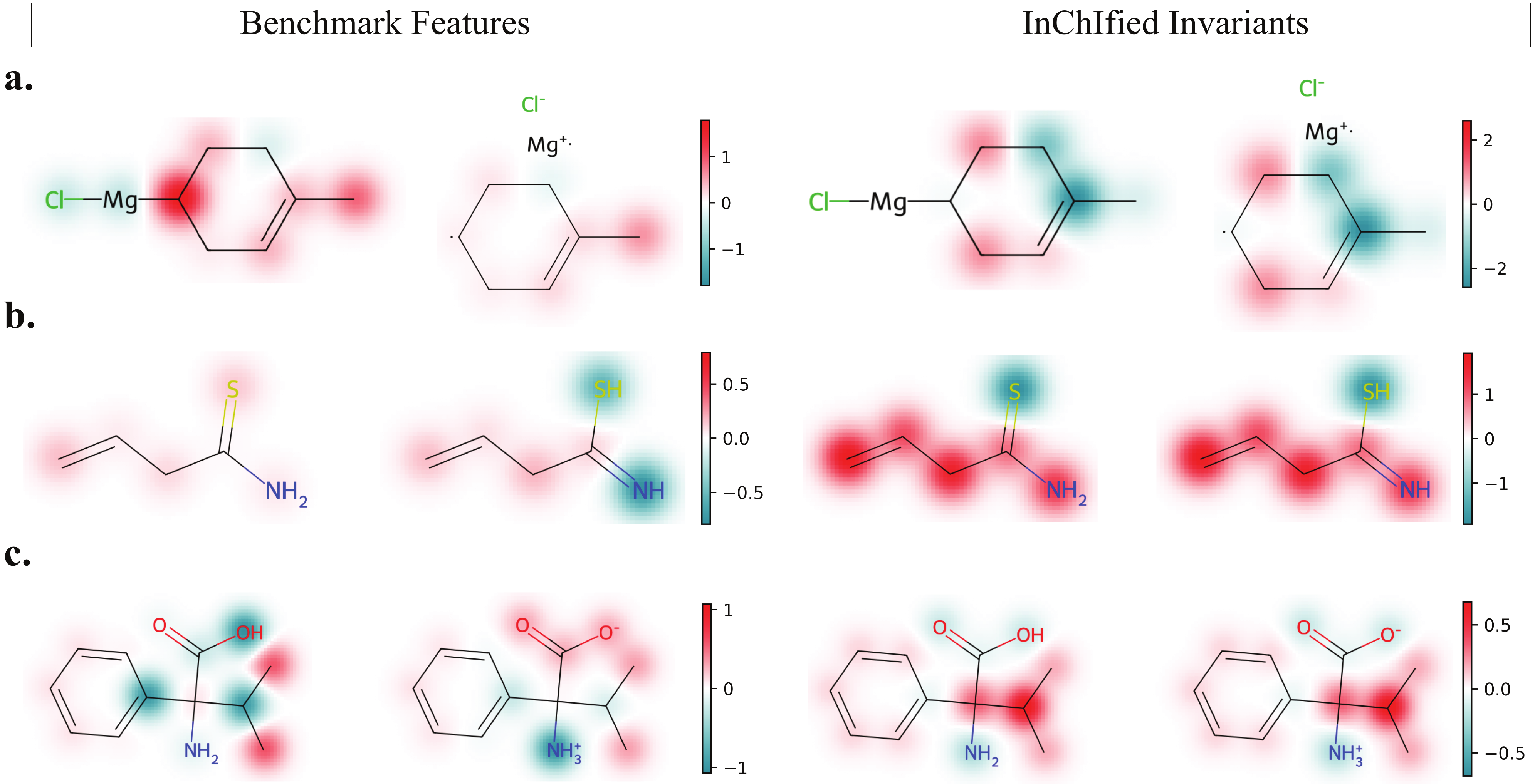}
    \caption{Attribution maps. Each panel reports attributions for two graphs with the same InChI generated by models trained with benchmark features (left) and \ii~(right). The models are neural graph fingerprints as described in \Cref{sec:performance}, and attributions are generated as described in \Cref{sec:explainability}. The colormap indicates the contribution of nodes to the final prediction for regression tasks and to the logits for classification tasks. The tasks are: \textbf{a.} BACE; \textbf{b.} BBBP; \textbf{c.} Lipo.}
    \label{fig:explainability}
\end{figure}

Figure~\ref{fig:explainability} reports three examples (\textbf{a.} metal disconnection; \textbf{b.} tautomerism; \textbf{c.} zwitterions) where two different input graphs with the same chemical identity yield divergent attributions when using standard featurization methods. For instance, this is evident on the carbon connected to \ch{Mg} atom that bears a radical electron after \ch{MgCl} disconnection (panel \textbf{a.}), on the two tautomers for the primary thioamide \ch{CSNH2} (panel \textbf{b.}), and on the carboxylic acid \ch{COOH} and primary amine \ch{NH2} moieties that can be written in a zwitterionic form (panel \textbf{c.}). \ii~resolve these inconsistencies and yield identical attribution maps across alternative graph representations with the same chemical identity, as they enforce explanation invariance by construction. 

Overall, this work contributes to the development of novel featurization methods for molecular machine learning, a direction advocated by recent benchmarking work as underexplored yet instrumental to advancing the field~\cite{liu2024welqrate}.

\paragraph{Contributions.}
Our contributions are as follows.
\begin{itemize}
    \item We introduce a new featurization method for machine learning on molecular graphs: \ii~are node, edge, and graph features designed to be invariant under transformations of the molecular graph that preserve chemical identity.
    \item We validate \ii~on a large dataset from PubChem: they produce identical molecular representations for graphs with the same InChI in more than $99\%$ of cases, whereas standard Daylight invariants do so in fewer than $1\%$ of cases.
    \item We evaluate predictive performance on MoleculeNet benchmark tasks: \ii~preserve predictive performance while improving consistency across alternative graph depictions with the same chemical identity.
    \item We perform a quantitative analysis of attribution consistency: attributions generated with standard features vary substantially across chemically equivalent graphs, whereas attributions generated with \ii~achieve almost perfect consistency.
    \item We release open-source software implementing \ii: they can be seamlessly integrated into existing workflows as a replacement for standard molecular graph features.
\end{itemize}

\paragraph{Related work.}
InChI was developed by IUPAC as a canonical identifier for molecular identity and is widely used for database indexing and structure curation~\cite{heller2015inchi,bento2020open}, but its use in machine learning has largely been limited to textual descriptors rather than node- and edge-level graph features~\cite{handsel2021translating,wojtuch2023extended,wang2025recent}. Recent work has also explored chemically motivated invariances in molecular representations: most closely, RIGR enforces resonance invariance by mapping different resonance forms of a molecule to a common graph representation~\cite{zalte2025rigr}, while data augmentation has been used to encourage selected invariances such as tautomeric invariance~\cite{ulrich2021exploring}. However, RIGR targets a specific class of Lewis-structure ambiguity, whereas augmentation-based approaches require generating multiple equivalent graph representations for each molecule, which can be computationally expensive and may be infeasible when the relevant equivalence class is large or difficult to enumerate. In contrast, \ii~use InChI to define the target equivalence relation and derive node-, edge-, and graph-level features reflecting a broader set of chemical standardization operations, including resonance/charge normalization, proton relocation, tautomerism, fragment handling, metal disconnection, and stereochemical conventions. Thus, \ii~provide a featurization framework aligned with an international chemical identity standard. Existing molecular graph models and fingerprints are typically invariant to atom ordering or graph isomorphism~\cite{rogers2010extended,duvenaud2015convolutional,huang2022going}, and some architectures incorporate equivariance or invariance to physical transformations~\cite{schutt2017schnet}, but these properties do not guarantee invariance across chemically equivalent graph depictions. Finally, explanation methods such as LIME, SHAP, Integrated Gradients, GNNExplainer, PGExplainer, and attention-based explanations identify features or subgraphs driving predictions for a fixed input representation~\cite{ribeiro2016should,lundberg2017unified,sundararajan2017axiomatic,ying2019gnnexplainer,luo2020parameterized,xiong2019pushing}. They do not, however, ensure that explanations remain consistent when different graph inputs represent the same molecule. An extended discussion of related work is provided in Appendix~\ref{sec:background}.

\section{Methodology}\label{sec:methodology}

Our goal is to obtain atom (node), bond (edge), and structural (graph) features that are the same for graphs with the same Standard InChI and that are different for graphs with different Standard InChIs. We call such features \ii. 

We use the following notation. A molecular structure is represented with a graph $\mathcal{G}(\mathcal{A}, \mathcal{B})$ composed of a set of atoms $\mathcal{A}$ and a set of bonds $\mathcal{B} \subseteq \mathcal{A} \times \mathcal{A}$. Each atom $\alpha_i \in \mathcal{A}$ is labeled with a vector of attributes $\mathbf{a}_i$ (Table~\ref{tab:notation:ai}). Each bond $\beta_{ij}=(\alpha_i, \alpha_j)\in \mathcal{B}$ is labeled with a vector of attributes $\mathbf{b}_{ij}$ (Table~\ref{tab:notation:bij}). A neighborhood function $\mathcal{N}(\alpha_i)=\{\alpha_j:(\alpha_j, \alpha_i)\in \mathcal{B}\}$ assigns the set of neighbors $\mathcal{N}(\alpha_i)$ to each atom $\alpha_i\in \mathcal{A}$. Given an input graph $\mathcal{G}$, we apply the transformations described in Steps 1 to 10 below. These steps are applied sequentially and constitute a single workflow inspired by the InChI Technical Manual~\cite{inchimanual}. Because the InChI code consists of approximately 112,000 lines of C code containing 970 functions and 130 macros in 37 source files~\cite{hull2011inchi}, these steps are designed to mimic the InChI standard as closely as possible while also making the process more concise, accessible, and interpretable. We provide illustrations for each step in \Cref{apx:figures}.

\paragraph{Step 1: Preparation.}
This step prepares the input graph for further processing. First, we calculate the total charge of the molecule $\texttt{Q}_\mathcal{A}$ by summing the charges of all the atoms $\alpha_i\in\mathcal{A}$ (Figure~\ref{fig:step_prepare_charge}). 
Second, aromatic bonds are converted into a series of single and double bonds using the aromaticity model and kekulization in RDKit~\cite{rdkit} (Figure~\ref{fig:step_prepare_kekulize}). 
Finally, hydrogen atoms are marked as phantom atoms and they are moved into the attributes of attached atoms (Figure~\ref{fig:step_prepare_hydrogens} and Algorithm~\ref{alg:hs}). Specifically, for each atom $\alpha_i \in \mathcal{A}$ that is a hydrogen atom ($\texttt{Z}_i = 1$), we check if the formal charge is positive and the isotope number is zero. If so, we mark the atom as a phantom atom. Then, for each neighbor $\alpha_j$ of $\alpha_i$, we check whether $\alpha_j$ is not a hydrogen atom or has an isotope number greater than or equal to that of $\alpha_i$. If this condition is met, we remove the bond between $\alpha_i$ and $\alpha_j$, increment the number of hydrogens attached to $\alpha_j$ by one unit, and mark $\alpha_i$ as a phantom atom. Additionally, if $\alpha_i$ is protium, deuterium, or tritium, we add it to $\texttt{SetHs}_j$, indicating the presence of the specific hydrogen isotope in the hydrogens attached to $\alpha_j$. We also add $\alpha_i$ to $\texttt{InitialSetHs}$, which stores the occurrence of hydrogen isotopes across all atoms in the molecule and facilitates the handling of mobile isotopes (Step~9 below). 

\paragraph{Step 2: Disconnection.}
This step disconnects metals by increasing the charge of the metal and decreasing the charge of the attached atoms (Figure~\ref{fig:step_disconnect} and Algorithm~\ref{alg:metals}). Metals are all atoms except \ch{H}, \ch{He}, \ch{B}, \ch{C}, \ch{N}, \ch{O}, \ch{F}, \ch{Ne}, \ch{Si}, \ch{P}, \ch{S}, \ch{Cl}, \ch{Ar}, \ch{Ge}, \ch{As}, \ch{Se}, \ch{Br}, \ch{Kr}, \ch{Te}, \ch{I}, \ch{Xe}, \ch{At}, \ch{Rn}~\cite{heller2015inchi}. For each metal atom $\alpha_i$, we drop all its attached hydrogens and its chiral tag. Then, for each neighbor $\alpha_j$ of $\alpha_i$, we remove the bond $\beta_{ij}$ between $\alpha_i$ and $\alpha_j$ and we update the formal charge of $\alpha_i$ by adding the order of the bond $\beta_{ij}$ and the number of radical electrons of $\alpha_j$. We update the formal charge of $\alpha_j$ by subtracting the same value and reset its number of radical electrons to 0. Finally, we drop the chiral tag of $\alpha_j$ and, for each neighbor $\alpha_k$ of $\alpha_j$, we drop the stereochemistry of the bond between $\alpha_j$ and $\alpha_k$.

\paragraph{Step 3: Normalization.}
This step normalizes the input by changing the distribution of formal charges and bond orders (Table~\ref{tab:step_normalize} and Figure~\ref{fig:step_normalize}). We use the SMILES arbitrary target specification (SMARTS) language to retrieve the molecular patterns listed in Table~\ref{tab:step_normalize} and apply the corresponding transformations. Each transformation is applied repeatedly, until no more matches are found, before passing to the next one. The stereochemistry of modified bonds is dropped.

\begin{table*}[t]
    \small
    \centering
    \caption{Sorted list of SMARTS patterns and corresponding transformations. Legend: \Z{N} = \ch{N}; \Z{C} = \ch{C}, \ch{O}, \ch{P}, \ch{S}; \Z{X} = \ch{C}, \ch{N}, \ch{O}, \ch{P}, \ch{S}, \ch{As}, \ch{Se}, \ch{Sb}, \ch{Te}, \ch{I}; $\Qp$ = positive charges; $\Qn$ = negative charges.}
    \label{tab:step_normalize}
    \begin{tabularx}{\textwidth}{c c X}
        \toprule
        \textsc{ID} & \textsc{SMARTS} & \textsc{Transformation} \\
        \midrule
        1 & $\underbrace{[\Z{X};\Qn]}_{i_0}-,=\underbrace{[\Z{X};\Qp]}_{i_1}$ & Increase the charge of $\alpha_{i_0}$ and decrease that of $\alpha_{i_1}$ by one unit. Increase the bond order of $\beta_{i_0i_1}$ by one unit. \\
        2 & $\underbrace{[\Z{X};\Qn]}_{i_0}-\underbrace{*}_{i_1}=\underbrace{[\Z{X};\Qp]}_{i_2}$ & Increase the charge of $\alpha_{i_0}$ and decrease that of $\alpha_{i_2}$ by one unit. Increase the bond order of $\beta_{i_0i_1}$ and decrease that of $\beta_{i_1i_2}$ by one unit.  \\
        3 & $\underbrace{[\Z{C};\Qp]}_{i_0}-\underbrace{[\Z{N};\text{!H0}]}_{i_1}$ & Decrease the charge of $\alpha_{i_0}$ and increase that of $\alpha_{i_1}$ by one unit. Increase the bond order of $\beta_{i_0i_1}$ by one unit. \\
        4 & $\underbrace{[\Z{C};\Qp]}_{i_0}=\underbrace{*}_{i_1}-\underbrace{[\Z{N}]}_{i_2}$ & Decrease the charge of $\alpha_{i_0}$ and increase that of $\alpha_{i_2}$ by one unit. Decrease the bond order of $\beta_{i_0i_1}$ and increase that of $\beta_{i_1i_2}$ by one unit.  \\
        5 & $\underbrace{[\Z{N};\Qp]}_{i_0}-\underbrace{*}_{i_1}=\underbrace{*}_{i_2}-\underbrace{[\Z{N};\Qn;\text{!H0}]}_{i_3}$ & Decrease the charge of $\alpha_{i_0}$ and increase that of $\alpha_{i_3}$ by one unit. Increase, decrease, and increase the bond order of $\beta_{i_0i_1}$, $\beta_{i_1i_2}$, and $\beta_{i_2i_3}$, respectively, by one unit. \\
        \bottomrule
    \end{tabularx}
\end{table*}

\paragraph{Step 4: Deprotonation.}
This step removes protons from heteroatoms as follows (Figure~\ref{fig:step_deprotonate}). First, we identify positively charged heteroatoms using the SMARTS in~\eqref{eq:soft}:
\begin{equation}\label{eq:soft}
    \underbrace{[\Z{Z};\text{!H0};\Qp]}_{i}
\end{equation}
where \Z{Z} = \ch{N}, \ch{O}, \ch{F}, \ch{P}, \ch{S}, \ch{Cl}, \ch{Se}, \ch{Br}, \ch{Te}, \ch{I} and $\Qp$ is one or more positive charges. We decrease the charge of $\alpha_i$ and its number of attached hydrogens by one unit, and we set $\texttt{ChiralTag}_i=0$ and $\texttt{SetHs}_i=\{\}$. This transformation is applied repeatedly until no more matches are found.
Second, we retrieve the molecular pattern described by the SMARTS in~\eqref{eq:hard}:
\begin{equation}\label{eq:hard}
    \underbrace{[\Z{X};\text{!H0}]}_{i_0}\underbrace{-*=*-*=*-...=}_{i_1,i_2,...,i_{2n+1}}\underbrace{[\Z{N};\Qp]}_{i_{2n+2}}
\end{equation}
where \Z{X} = \ch{N}, \ch{O}, \ch{S}, \ch{Se}, \ch{Te}, \Z{N} = \ch{N}, $\Qp$ is one or more positive charges, and $n$ is a non-negative integer that controls the length of the chain. By default, we use $n=0,1,2,3,4$. We decrease the charge of $\alpha_{i_{2n+2}}$ by one unit, we decrease the number of hydrogens attached to $\alpha_{i_0}$ by one unit, and we set $\texttt{ChiralTag}_{i_0}=0$ and $\texttt{SetHs}_{i_0}=\{\}$. Single bonds are converted to double bonds and vice versa and their stereochemistry information is dropped. This transformation is applied repeatedly until no more matches are found.

\paragraph{Step 5: Neutralization.}
This step attempts to neutralize each fragment in the molecule (Figure~\ref{fig:step_neutralize}). For each fragment $\mathcal{F}\subseteq\mathcal{A}$, we calculate its total charge $\texttt{Q}_{\mathcal{F}}$ by summing the charges of the atoms $\alpha_i\in\mathcal{F}$. 
If $\texttt{Q}_{\mathcal{F}}$ is positive, we search $\mathcal{F}$ for the pattern given by the SMARTS in~\eqref{eq:pos}:
\begin{equation}\label{eq:pos}
    \underbrace{[\Z{X};\text{H1}]}_{i}
\end{equation}
where \Z{X} = \ch{O}, \ch{F}, \ch{P}, \ch{S}, \ch{Cl}, \ch{Br}, \ch{I}. We decrease the charge of $\alpha_i$ and its number of attached hydrogens by one unit, set $\texttt{ChiralTag}_i=0$ and $\texttt{SetHs}_i=\{\}$. This process is applied repeatedly until the fragment has a charge of zero or no more matches are found. 
If $\texttt{Q}_{\mathcal{F}}$ is negative, we search $\mathcal{F}$ for the pattern given by the SMARTS in~\eqref{eq:neg}:
\begin{equation}\label{eq:neg}
    \underbrace{[\Z{X};\Qn]}_{i}
\end{equation}
where \Z{X} = \ch{O}, \ch{F}, \ch{P}, \ch{S}, \ch{Cl}, \ch{Br}, \ch{I}, and $\Qn$ is one or more negative charges. We increase the formal charge of $\alpha_i$ and its number of attached hydrogens by one unit, and set $\texttt{ChiralTag}_i=0$. This process is applied repeatedly until the fragment has a charge of zero or no more matches are found. 

\paragraph{Step 6: Valence reduction.}
This step drops \ch{H2} from non-metal atoms to reduce their valence (Figure~\ref{fig:step_drop_h2}). First, we identify \ch{B}, \ch{C}, \ch{N}, \ch{O}, \ch{F}, \ch{Si}, \ch{P}, \ch{S}, \ch{Cl}, \ch{Ge}, \ch{As}, \ch{Se}, \ch{Br}, \ch{Te}, \ch{I}, \ch{At} atoms that are bound to at least two hydrogens. Then, we drop the two hydrogens if the atom has a valence equal or greater than its lowest standard valence plus two. This process is applied repeatedly and we set $\texttt{SetHs}_i=\{\}$ and $\texttt{ChiralTag}_i=0$ for all atoms $\alpha_i$ that have been modified. We use the standard valences defined in \cite{inchimanual}.

\paragraph{Step 7: Movable charge detection.}
This step identifies alternating bond chains that connect positive charges located on nitrogen atoms using the SMARTS in~\eqref{eq:step_mobile_charge}:
\begin{equation}\label{eq:step_mobile_charge}
    \underbrace{[\Z{N};\text{H0};+1]}_{i_0}\underbrace{=*-*=...-}_{i_1,i_2,...,i_{2n+1}}\underbrace{[\Z{N};\text{H0}]}_{i_{2n+2}}
\end{equation}
where \Z{N} = \ch{N} and $n$ is a non-negative integer that controls the length of the chain. By default, we use $n=0,1,2,3,4$. These bonds are tagged as ``aromatic'' and their stereochemistry information is dropped (Figure~\ref{fig:step_mobile_charge}). The tag ``aromatic'' is not to be intended as the chemical definition of aromaticity, and it is solely used to identify alternating bonds in the next step.

\paragraph{Step 8: Tautomerism detection.}
This step detects tautomeric patterns in the molecule and removes mobile hydrogens (Figure~\ref{fig:step_mobile_hs} and Algorithm~\ref{alg:tautomerism}). We search iteratively for tautomeric patterns by initializing an empty set $\mathcal{T}$ and repeating the following steps until the set $\mathcal{T}$ remains unchanged between two subsequent iterations. For each iteration, we identify tautomeric patterns using the SMARTS in~\eqref{eq:tautomerism}:
\begin{equation}\label{eq:tautomerism}
    \underbrace{[\Z{M}]}_{i_0}\underbrace{=,:[\Z{Q}]-,:[\Z{Q}]=,:...-,:}_{i_1,i_2,...,i_{2n+1}}\underbrace{[\Z{M};\text{!H0},-]}_{i_{2n+2}}
\end{equation}
where \Z{M} = \ch{N}, \ch{O}, \ch{S}, \ch{Se}, \ch{Te}, \Z{Q} = \ch{C}, \ch{N}, \ch{S}, \ch{P}, \ch{Sb}, \ch{As}, \ch{Se}, \ch{Te}, \ch{Br}, \ch{Cl}, \ch{I}, and $n$ is a non-negative integer that controls the length of the chain. By default, we use $n=0,1,2,3,4$. When this pattern is matched, we add $\alpha_{i_0}$ and $\alpha_{i_{2n+2}}$ to the set $\mathcal{T}$. We set the formal charges of $\alpha_{i_0}$ and $\alpha_{i_{2n+2}}$ to $-1$, drop the hydrogens attached to the two atoms, and set $\texttt{SetHs}_{i_0}=\texttt{SetHs}_{i_{2n+2}}=\{\}$ and $\texttt{ChiralTag}_{i_0}=\texttt{ChiralTag}_{i_{2n+2}}=0$. The stereochemistry is dropped from all bonds belonging to $\alpha_{i_0}$ or $\alpha_{i_{2n+2}}$. Finally, we drop the stereochemistry from all bonds in the chain and tag them as ``aromatic'' to enable the discovery of new tautomeric patterns in the next iteration.

\paragraph{Step 9: Hydrogen isotopes.}
This step removes the phantom tag from mobile protiums, deuteriums, and tritiums (Figure~\ref{fig:step_isotopes_hs}). Specifically, we set $\texttt{Phantom}_i=\text{False}$ for each atom $\alpha_i$ that is in the set $\texttt{InitialSetHs}$ but not in the union of the sets $\cup_{\alpha_i\in\mathcal{A}} \texttt{SetHs}_i$. Moreover, for each atom $\alpha_j\in\mathcal{A}$, we set the tags $\texttt{Num1H}_j$, $\texttt{Num2H}_j$, and $\texttt{Num3H}_j$ equal to the number of elements $\alpha_k \in \texttt{SetHs}_j$ such that $\texttt{Isotope}_k=1,2,3$, respectively.

\paragraph{Step 10: Stereochemistry.}
This step assigns $R/S$ and $E/Z$ stereochemical labels based on Cahn--Ingold--Prelog (CIP) rules as described in~\cite{hanson2018algorithmic}. The $R/S$ and $E/Z$ labels for the atom $\alpha_i$ and bond $\beta_{ij}$ are stored in the variables $\texttt{CIPCode}_i$ and $\texttt{BondCIPCode}_{ij}$, respectively. We drop the label in those cases that are not treated as possibly stereogenic by InChI~\cite{inchimanual} (Figure~\ref{fig:step_stereochemistry}). Specifically, we set $\texttt{BondCIPCode}_{ij}=0$ for bonds where at least one of the two connected atoms is not \ch{C}, \ch{N}, \ch{Si}, \ch{Ge}, and we set $\texttt{CIPCode}_i=0$ for atoms that meet one or more of the following conditions: (i) the atom is not \ch{B}, \ch{C}, \ch{N}, \ch{Si}, \ch{P}, \ch{S}, \ch{Ge}, \ch{As}, \ch{Se}, \ch{Sn}; (ii) the atom is \ch{N} with a number of neighbours different from 4 and it is not in a ring of size 3; (iii) the atom is \ch{N}, \ch{P}, \ch{As}, \ch{S}, \ch{Se}, with at least one attached hydrogen or two terminal neighbors \Z{X}\ch{H_n} and \Z{X}\ch{H_m}, with $n+m>0$ where \Z{X} $=$ \ch{O}, \ch{S}, \ch{Se}, \ch{Te}, \ch{N}.

Finally, \ii~are defined as follows. The atom (node) invariants are the attributes \texttt{Z}, \texttt{Isotope}, \texttt{Degree}, \texttt{InRing}, \texttt{NumHs}, \texttt{Num1H}, \texttt{Num2H}, \texttt{Num3H}, and \texttt{CIPCode} that are obtained after applying the transformations from Step 1 to 10 above. Each atom also has a tag \texttt{Phantom} indicating whether it must be ignored. A phantom atom has no bonds to any other atom. The reason why we do not drop it directly is to maintain the original atom indices, which greatly simplifies, for instance, mapping explanations back to the original input graph for visualization. For bonds (edges), the only invariant is \texttt{BondCIPCode}. Notice that some bonds may have been dropped during the transformations, and thus only the remaining bonds should be considered. Finally, we also include the total charge $\texttt{Q}_\mathcal{A}$ as a graph invariant.

\section{Experiments}\label{sec:experiments}

This section presents our experiments. All results have been generated using Python 3.10 on macOS Sonoma with M2~chip and 16~GB memory. We proceed as follows. First, we validate the proposed \ii~(\Cref{sec:validation}). Then, we assess the predictive performance (\Cref{sec:performance}). Finally, we provide a quantitative analysis of attribution consistency (\Cref{sec:explainability}).

\subsection{Validation: InChIfied invariants align featurization with chemical identity}\label{sec:validation}

We download the first 477,500,000 substances in Structured-Data Format (SDF) from PubChem Substances~\cite{kim2016pubchem,kim2025pubchem}, a public repository maintained by the National Center for Biotechnology Information (NCBI) under the United States National Institutes of Health (NIH). This repository collects chemical substance data provided directly by a diverse range of contributors, including laboratories, chemical vendors, pharmaceutical companies, and researchers. For each substance, we generate the corresponding canonical SMILES string and InChI Key. Substances with more than 100 atoms or that cannot produce a valid SMILES string or InChI Key are ignored. The final dataset contains 106,297,689 unique (SMILES, InChI Key) pairs. We count the number of SMILES strings for each  InChI Key and sort the pairs such that InChI Keys associated with the highest number of SMILES strings come first. Alphabetical ordering of InChI Keys is used to break ties. Then, we select the first million SMILES strings and the associated InChI Keys. 

\begin{table}[t]
    \centering
    \caption{Confusion matrices counting the number of times where fingerprints $f$ match depending on whether InChI Keys $k$ match. \textbf{a.} With \ii; \textbf{b.} With benchmark features.}
    \label{tab:confusion}
    \begin{subtable}[t]{.47\textwidth}
    \begin{tabularx}{\textwidth}{Xrr}
        \toprule
        \textbf{a.} InChIfied & $f_i=f_j$ & $f_i\neq f_j$ \\
        \toprule
        $k_i=k_j$ & 911,534 & 3,452 \\
        $k_i\neq k_j$ & 13,199 & 98,065,083 \\
        \bottomrule
    \end{tabularx}
    \end{subtable}
    \hspace{1em}
    \begin{subtable}[t]{.47\textwidth}
    \begin{tabularx}{\textwidth}{Xrr}
        \toprule
        \textbf{b.} Benchmark & $f_i=f_j$ & $f_i\neq f_j$ \\
        \toprule
        $k_i=k_j$ & 3,241 & 911,745 \\
        $k_i\neq k_j$ & 335 & 98,077,947 \\
        \bottomrule
    \end{tabularx}
    \end{subtable}
\end{table}

\paragraph{Representation consistency.} 
For each SMILES string in the dataset, we compute the corresponding Extended-Connectivity Fingerprints (ECFP) with radius 2 using either the standard Daylight invariants in the original paper~\cite{rogers2010extended} or the \ii. We use sparse count fingerprints, not fixed-length hashed fingerprints, to avoid collisions induced by finite fingerprint size. For \ii, we reserve keys 0 and 1 to encode the total charge $\texttt{Q}_\mathcal{A}$ (if $\texttt{Q}_\mathcal{A}<0$ then $-\texttt{Q}_\mathcal{A}$ is used for the count of key 0, if $\texttt{Q}_\mathcal{A}>0$ then $\texttt{Q}_\mathcal{A}$ is used for the count of key 1). We compare each InChI Key with the next 100 InChI Keys, and test whether the resulting sparse count fingerprints are identical when InChI Keys are equal, and they are different when InChI Keys are different. Because the data are sorted by InChIKey and no key has more than 100 SMILES, this captures all positive matches while avoiding infeasible (roughly $10^{12}/2$) all-pairs comparisons. Table~\ref{tab:confusion} reports the results. When InChI Keys differ, both fingerprints differ in more than 99.98\% of cases (see Appendix~\ref{apx:keto} for further information). When InChI Keys match, fingerprints derived from the \ii~also match in 99.62\% of cases, whereas those derived from the standard Daylight invariants match in only 0.35\% of cases.

\paragraph{Similarity across equivalent graphs.}
\begin{wraptable}{r}{0.43\linewidth}
  \caption{Quantiles of the generalized Tanimoto coefficient between fingerprints of InChI-equivalent graph pairs, computed using standard Daylight invariants at different fingerprint radii.}
  \label{tab:radius_quantile_results}
  \centering
  \begin{tabular}{lccc}
    \toprule
     & \multicolumn{3}{c}{\textsc{Radius}} \\
    \cmidrule(lr){2-4}
    \textsc{Quantile} & \textsc{2} & \textsc{4} & \textsc{6} \\
    \midrule
    \textsc{10\%} & 0.63 & 0.51 & 0.49 \\
    \textsc{25\%} & 0.77 & 0.63 & 0.59 \\
    \textsc{50\%} & 0.86 & 0.74 & 0.68 \\
    \textsc{75\%} & 0.91 & 0.81 & 0.75 \\
    \textsc{90\%} & 0.94 & 0.87 & 0.82 \\
    \bottomrule
  \end{tabular}
\end{wraptable}
We additionally quantified fingerprint similarity for InChI-equivalent pairs using the generalized Tanimoto coefficient. Quantiles of the Tanimoto coefficient distribution at different fingerprint radii using standard Daylight invariants are reported in \Cref{tab:radius_quantile_results}. Tanimoto values in the 0.5--0.7 range indicate that the corresponding fingerprints can differ substantially, so these discrepancies are not limited to minor perturbations. Moreover, similarity decreases as the fingerprint radius increases. This suggests that representation inconsistencies can be amplified as information is aggregated over larger molecular neighborhoods, and that more expressive models may therefore be more sensitive to such inconsistencies. For fingerprints derived from the \ii, all corresponding quantiles are exactly 1 and are therefore omitted from the table.

\subsection{Performance: InChIfied invariants preserve predictive performance}\label{sec:performance}

We train and evaluate all models on ten MoleculeNet benchmark datasets~\cite{wu2018moleculenet}. To evaluate how models behave across alternative but equivalent graph depictions, we additionally construct datasets containing graphs that differ from the original graphs but correspond to the same chemical identity (see \Cref{sec:augmentation}). We refer to them as alternative-depiction datasets.

\paragraph{Input representation.}
The \ii~are used as a drop-in replacement for the benchmark features. Atom- and bond-level \ii~are one-hot encoded in the same way as the corresponding benchmark features. The graph-level total charge is appended to the atom features, and phantom atoms are retained in the graph with zero-valued features for convenience (see \Cref{sec:phantom} for details). No architectural changes are made.

\paragraph{Models and training.}
As a baseline model, we use neural graph fingerprints with a linear layer on top~\cite{duvenaud2015convolutional}, as they represent a seminal work in deep learning for molecular graphs. We evaluate this model with either the benchmark features defined in the original work or the \ii. We set the fingerprint radius to 3, the size of the hidden layers to 128, and the fingerprint length to 64. We train the model using the Adam optimizer~\cite{kingma2014adam} with a learning rate of 0.001 and a batch size of 100 for 500 epochs. These hyperparameters are chosen based on \cite{duvenaud2015convolutional}. 
We also evaluate attentive fingerprints~\cite{xiong2019pushing}, again using either the corresponding benchmark features or the \ii, while keeping the remaining hyperparameters as in the original paper.
The loss is the mean squared error for regression tasks and the cross-entropy loss for classification tasks.  We randomly split the data into train (80\%), validation (10\%), and test (10\%) sets. We train on the original training split and select the checkpoint with the best validation score on the original validation split. We then evaluate the selected checkpoint both on the original test split and on the corresponding alternative-depiction dataset constructed from the full original dataset. The score is the root mean square error (RMSE) for regressions and the ROC-AUC score for classifications. We repeat each experiment five times and report the mean test score with its standard error.

\paragraph{Predictive performance.}
On the original MoleculeNet test sets, models using \ii~perform comparably to models using the corresponding benchmark features (see~\Cref{tab:moleculenet} in the appendix). This shows that replacing benchmark features with \ii~preserves predictive performance under the standard evaluation protocol. The difference appears on the alternative-depiction datasets: as shown in \Cref{tab:moleculenet_new_smiles}, models using \ii~perform significantly better than models using benchmark features across nearly all datasets and both architectures. If changing the graph depiction induced only a negligible perturbation under benchmark features, we would expect the two featurizations to perform similarly on these alternative depictions as well. The observed gap therefore indicates that benchmark features can make predictions sensitive to graph depiction, whereas \ii~significantly reduce this sensitivity.

\begin{table*}
  \caption{Scores and standard errors of neural fingerprints (NeuralFP) and attentive fingerprints (AttentiveFP) on the alternative-depiction datasets, using the features in the original papers (Benchmark) and \ii~(InChIfied). Bold values indicate scores that are statistically better between Benchmark and InChIfied models using a two-sided $z$-test at the 95\% confidence level.}
  \label{tab:moleculenet_new_smiles}
  \centering
  \begin{tabular}{lc|cc|cc}
    \toprule
    \textsc{Dataset} & \textsc{Score} & \textsc{NeuralFP} & \textsc{NeuralFP} & \textsc{AttentiveFP} & \textsc{AttentiveFP} \\
     & & \textsc{(Benchmark)} & \textsc{(InChIfied)} & \textsc{(Benchmark)} & \textsc{(InChIfied)} \\
    \midrule
    ESOL     & RMSE    & 0.83 $\pm$ 0.10 & 0.70 $\pm$ 0.05 & 0.59 $\pm$ 0.04 & \textbf{0.35 $\pm$ 0.04} \\
    FreeSolv & RMSE    & 2.74 $\pm$ 0.08 & \textbf{1.62 $\pm$ 0.24} & 3.56 $\pm$ 0.46 & \textbf{0.97 $\pm$ 0.23} \\
    Lipo     & RMSE    & 1.29 $\pm$ 0.06 & \textbf{0.51 $\pm$ 0.02} & 1.23 $\pm$ 0.11 & \textbf{0.43 $\pm$ 0.02} \\
    HIV      & ROC-AUC & 0.88 $\pm$ 0.01 & \textbf{0.94 $\pm$ 0.01} & 0.91 $\pm$ 0.01 & 0.93 $\pm$ 0.02 \\
    BACE     & ROC-AUC & 0.91 $\pm$ 0.01 & \textbf{0.95 $\pm$ 0.00} & 0.84 $\pm$ 0.02 & \textbf{0.91 $\pm$ 0.00} \\
    BBBP     & ROC-AUC & 0.91 $\pm$ 0.01 & \textbf{0.95 $\pm$ 0.01} & 0.92 $\pm$ 0.01 & \textbf{0.94 $\pm$ 0.00} \\
    Tox21    & ROC-AUC & 0.77 $\pm$ 0.01 & \textbf{0.87 $\pm$ 0.01} & 0.82 $\pm$ 0.02 & \textbf{0.89 $\pm$ 0.01} \\
    ToxCast  & ROC-AUC & 0.78 $\pm$ 0.01 & \textbf{0.83 $\pm$ 0.01} & 0.84 $\pm$ 0.02 & 0.88 $\pm$ 0.01 \\
    SIDER    & ROC-AUC & 0.66 $\pm$ 0.02 & 0.71 $\pm$ 0.02 & 0.75 $\pm$ 0.01 & 0.78 $\pm$ 0.03 \\
    ClinTox  & ROC-AUC & 0.81 $\pm$ 0.01 & \textbf{0.90 $\pm$ 0.01} & 0.86 $\pm$ 0.01 & \textbf{0.91 $\pm$ 0.01} \\
    \bottomrule
  \end{tabular}
\end{table*}

\subsection{Explainability: InChIfied invariants improve explanation consistency}\label{sec:explainability}

Models using \ii~receive the same input representation for chemically equivalent graphs in more than 99\% of cases (\Cref{sec:validation}) and therefore yield the same predictions and attributions by construction. The purpose of the present analysis is therefore to quantify the variability under benchmark features.

\paragraph{Prediction variability.}
We use the trained models to predict the 1 million SMILES from the PubChem dataset introduced in Section~\ref{sec:validation}. For simplicity, we restrict our focus to classification tasks, as the predicted probabilities are consistently scaled between 0 and 1, allowing for meaningful comparisons. We omit the ToxCast model because it contains over 600 individual tasks and complicates the visualization. For each of the remaining tasks, we group the SMILES strings that correspond to the same InChI, and within each group, we compute the difference between the maximum and minimum predicted probabilities. Figure~\ref{fig:boxplot} reports the distribution of these differences. The variation is not uniform across tasks, suggesting task-dependent sensitivity to input representations. The observed differences can be substantial. In particular, for the HIV task, more than 10\% of InChIs show a difference greater than 30\% in predicted probability depending on the specific SMILES with which they are represented. Additionally, for HIV, BACE, and BBBP, we find that the classification label (based on a 50\% probability threshold) is ambiguous for approximately 10\% to 11\% of InChIs. These results indicate that prediction variability is not negligible under benchmark features. 

\paragraph{Attribution variability.}
We sample 10,000 random pairs of SMILES strings with the same InChIKey from the PubChem dataset and generate attributions using Integrated Gradients~\cite{sundararajan2017axiomatic}, which has been found to be among the most faithful attribution methods for molecular graphs~\cite{sanchez2020evaluating, agarwal2023evaluating}. We use feature-level masking and sum feature attributions over each atom to obtain node-level attribution scores. Because equivalent graphs may differ in node order and number of nodes, we sort nodes according to their attribution weights under the InChIfied model and ignore phantom atoms. We then compare the paired attributions using Kendall's tau, as molecular explanations are typically interpreted in terms of relative atom importance rather than absolute attribution magnitude. We perform this analysis using the models trained on ClinTox and HIV, which correspond to low- and high-dispersion settings in \Cref{fig:boxplot}. Quantiles of the resulting Kendall's tau distributions are reported in \Cref{tab:quantile_results}. With benchmark features, attribution differences are non-negligible even in the more favorable ClinTox setting and widen substantially on HIV. The decrease in agreement is larger for attentive fingerprints than for neural fingerprints, consistent with the idea that architectures aggregating richer molecular context may be more sensitive to representation inconsistencies.

\begin{table*}
  \caption{Quantiles of Kendall's tau between attributions for pairs of SMILES strings with the same InChIKey, sampled from the PubChem dataset. Attributions are generated using neural fingerprints (NeuralFP) and attentive fingerprints (AttentiveFP) trained on the ClinTox and HIV datasets with the features in the original papers (Benchmark) and \ii~(InChIfied).}
  \label{tab:quantile_results}
  \centering
  \begin{tabular}{lllccccc}
    \toprule
    \textsc{Dataset} & \textsc{Model} & \textsc{Invariants} & \textsc{10\% Q} & \textsc{25\% Q} & \textsc{50\% Q} & \textsc{75\% Q} & \textsc{90\% Q} \\
    \midrule
    ClinTox & NeuralFP    & Benchmark  & 0.86 & 0.92 & 0.96 & 0.99 & 1.00 \\
            &             & InChIfied & 1.00 & 1.00 & 1.00 & 1.00 & 1.00 \\
            & AttentiveFP & Benchmark  & 0.71 & 0.91 & 0.95 & 0.98 & 0.99 \\
            &             & InChIfied & 1.00 & 1.00 & 1.00 & 1.00 & 1.00 \\
    HIV     & NeuralFP    & Benchmark  & 0.47 & 0.64 & 0.78 & 0.91 & 1.00 \\
            &             & InChIfied & 1.00 & 1.00 & 1.00 & 1.00 & 1.00 \\
            & AttentiveFP & Benchmark  & 0.35 & 0.52 & 0.68 & 0.84 & 0.95 \\
            &             & InChIfied & 0.99 & 1.00 & 1.00 & 1.00 & 1.00 \\
    \bottomrule
  \end{tabular}
\end{table*}

\paragraph{Final remarks.} We remark that all results have been obtained under a scenario where the fingerprints are differentiable, the baseline model is relatively simple, and the attribution method is considered faithful. We expect the issue of different explanations to become even more pronounced when using less faithful attribution methods, more complex models, and hash-based fingerprints, as they are more sensitive to small input perturbations and can lead to larger discrepancies. \ii~should be particularly useful in such contexts.

\section{Limitations}\label{sec:limitations}

\paragraph{Computational overhead.}
\ii~are more expensive to compute than standard Daylight invariants. However, this cost is typically paid only once during preprocessing, remains below the millisecond scale for an average molecule in our experiments, and is usually negligible relative to model training, inference, and explanation generation (see \Cref{tab:runtime_by_atoms}).

\paragraph{Scope of Standard InChI.}
\ii~inherit the notion of chemical identity encoded by Standard InChI. Some tasks may require optional or non-standard InChI layers to preserve additional distinctions. Nevertheless, Standard InChI is the relevant default for many molecular learning settings, because public chemical datasets are often curated, integrated, and deduplicated using Standard InChI or InChIKey upstream~\cite{goodman2021inchi,gaulton2012chembl}. In these cases, \ii~ align the model input with the same notion of chemical identity already used to construct the data.

\paragraph{Consistency is not explanation quality.}
\ii~improve explanation consistency for different graphs representing the same molecule, which is a prerequisite for explanations to be chemically meaningful, even though it is not by itself sufficient to establish explanation quality in general. Given that there is currently no widely accepted ground-truth benchmark for explanation quality in molecular machine learning that would enable a definitive large-scale evaluation of chemical meaningfulness, we view our contribution as a first step in that direction.

\section{Conclusion}\label{sec:conclusion}

The goal of \ii~is not to replace InChI for database lookup, but to bring the same chemical standard into graph-based machine learning as usable node, edge, and graph features. The approach inherits the scope of Standard InChI and may need to be extended when a task requires finer distinctions, such as non-standard InChI layers, detailed metal coordination, or full 3D structure. We release an open-source implementation of \ii, designed to be integrated into existing machine learning pipelines as a drop-in replacement for standard molecular graph features. Overall, \ii~provide a practical way to make representations, predictions, and explanations in molecular machine learning more consistent with chemical identity without requiring changes to downstream model architectures or explanation methods.

\bibliography{references}
\bibliographystyle{unsrt}

\newpage
\appendix
\onecolumn

\section{Supplementary algorithms}\label{sec:algorithms}
\renewcommand{\thealgorithm}{A.\arabic{algorithm}}
\setcounter{algorithm}{0}

\begin{minipage}[t]{0.50\textwidth}
\small
\begin{algorithm}[H]
\caption{Phantom hydrogen atoms.}
\label{alg:hs}
\begin{algorithmic}[1]
\For{$\alpha_i \in \mathcal{A}$}
    \If{$\texttt{Z}_i = 1$}
        \If{$\texttt{Q}_i > 0$ \textbf{and} $\texttt{Isotope}_i = 0$}
            \State $\texttt{Phantom}_i \gets \text{True}$
        \EndIf
        \For{$\alpha_j \in \mathcal{N}(\alpha_i)$} 
            \If{
                \Statex\quad\quad\quad\quad\quad
                $\texttt{Z}_j \neq 1$ \textbf{or}
                \Statex\quad\quad\quad\quad\quad 
                $\texttt{Isotope}_j \geq \texttt{Isotope}_i$
                \Statex\quad\quad\quad\quad\,
            }
                \State $\text{Drop bond } \beta_{ij}=(\alpha_i, \alpha_j)$
                \State $\texttt{NumHs}_j \mathrel{+}= 1$
                \State $\texttt{Phantom}_i \gets \text{True}$
                \If{$1\leq \texttt{Isotope}_i \leq 3$}
                    \State $\text{Add } \alpha_i \text{ to } \texttt{SetHs}_j$
                    \State $\text{Add } \alpha_i \text{ to } \texttt{InitialSetHs}$
                \EndIf
            \EndIf
        \EndFor
    \EndIf
\EndFor
\end{algorithmic}
\end{algorithm}
\end{minipage}
\hfill
\begin{minipage}[t]{0.48\textwidth}
\small
\begin{algorithm}[H]
\caption{Disconnection.}
\label{alg:metals}
\begin{algorithmic}[1]
\State $\mathcal{M}^\mathsf{c} = \{1,2,5,6,7,8,9,10,14,15,16,17,$
\Statex \quad\;$18,32,33,34,35,36,52,53,54,85,86\}$
\For{$\alpha_i \in \mathcal{A}$}
    \If{$\texttt{Z}_i \notin \mathcal{M}^\mathsf{c}$}
        \State $\texttt{NumHs}_i\gets 0$
        \State $\texttt{SetHs}_i\gets \{\}$
        \State $\texttt{ChiralTag}_i\gets 0$
        \For{$\alpha_j \in \mathcal{N}(\alpha_i)$} 
            \State $\text{Drop bond } \beta_{ij}=(\alpha_i, \alpha_j) $
            \State $\texttt{Q}_i \mathrel{+}= \texttt{BondOrder}_{ij} + \texttt{NumRs}_j$
            \State $\texttt{Q}_j \mathrel{-}= \texttt{BondOrder}_{ij} + \texttt{NumRs}_j$
            \State $\texttt{NumRs}_j \gets 0$
            \State $\texttt{ChiralTag}_j \gets 0$
            \For{$\alpha_k \in \mathcal{N}(\alpha_j)$} 
                \State $\texttt{BondStereo}_{jk} \gets 0$
            \EndFor
        \EndFor
    \EndIf
\EndFor
\end{algorithmic}
\end{algorithm}
\end{minipage}

\begin{algorithm}[H]
\caption{Tautomerism detection.}
\label{alg:tautomerism}
\begin{algorithmic}[1]
\State $\mathcal{T} = \{\}$, $\;t = -1$
\While{$|\mathcal{T}| \neq t$}
    \State $t \gets |\mathcal{T}|$
    \For{$n = 0,1,...$}
        \For{$(i_0,i_1,...,i_{2n+2})$ matching Equation~\eqref{eq:tautomerism}}
            \State Add $\alpha_{i_0}$ and $\alpha_{i_{2n+2}}$ to $\mathcal{T}$
            \State $\texttt{Q}_{i_0}\gets \texttt{Q}_{i_{2n+2}}\gets -1$
            \State $\texttt{NumHs}_{i_0}\gets\texttt{NumHs}_{i_{2n+2}}\gets0$
            \State $\texttt{SetHs}_{i_0}\gets\texttt{SetHs}_{i_{2n+2}}\gets\{\}$
            \State $\texttt{ChiralTag}_{i_0}\gets\texttt{ChiralTag}_{i_{2n+2}}\gets0$
            \For{$\beta_{ij}\in \{(\alpha_{i_0}, \alpha_j): \alpha_j \in \mathcal{N}(\alpha_{i_0})\} \cup \{(\alpha_{i_{2n+2}}, \alpha_j): \alpha_j \in \mathcal{N}(\alpha_{i_{2n+2}})\}$}
                \State $\texttt{BondStereo}_{ij} \gets 0$
            \EndFor
            \For{$\beta_{ij} \in \{(\alpha_{i_0}, \alpha_{i_1}),(\alpha_{i_1}, \alpha_{i_2}), ..., (\alpha_{i_{2n+1}}, \alpha_{i_{2n+2}})\}$}
                \State $\texttt{BondStereo}_{ij} \gets 0$
                \State $\texttt{BondOrder}_{ij} \gets 1.5$
            \EndFor
        \EndFor
    \EndFor
\EndWhile
\end{algorithmic}
\end{algorithm}

\section{Supplementary figures}\label{apx:figures}
\renewcommand{\thefigure}{B.\arabic{figure}}
\setcounter{figure}{0}

\begin{figure}[H]
    \centering
    \includesvg[width=\textwidth]{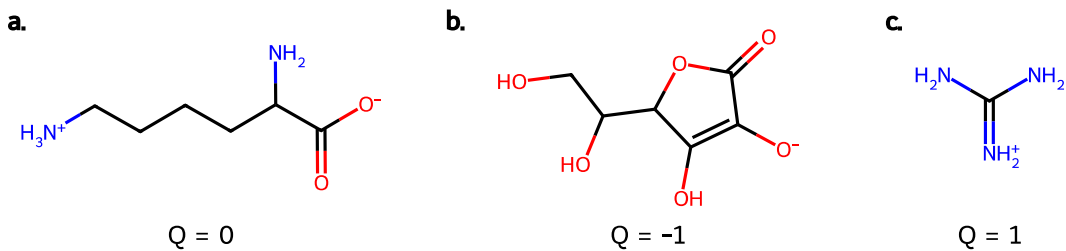}
    \caption{Illustration of Step~1: total charge. Computed total charge ($\texttt{Q}_\mathcal{A}$) for \textbf{a.}~a~neutral molecule ($\texttt{Q}_\mathcal{A}=0$); \textbf{b.}~a~negatively charged molecule ($\texttt{Q}_\mathcal{A}=-1$); \textbf{c.}~a~positively charged molecule ($\texttt{Q}_\mathcal{A}=1$).}
    \label{fig:step_prepare_charge}
\end{figure}

\begin{figure}[H]
    \centering
    \includesvg[width=\textwidth]{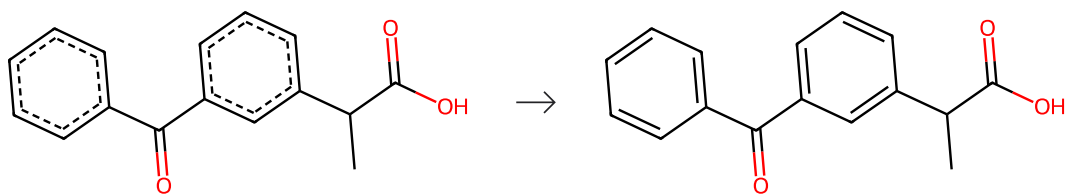}
    \caption{Illustration of Step 1: kekulization. Aromatic bonds are converted into a series of single and double bonds. The transformation is indicated by the arrow.}
    \label{fig:step_prepare_kekulize}
\end{figure}

\begin{figure}[H]
    \centering
    \includesvg[width=\textwidth]{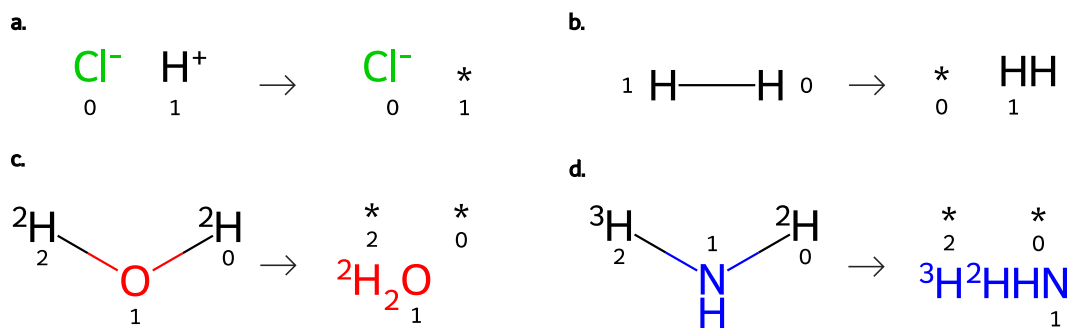}
    \caption{Illustration of Step 1: phantom hydrogen atoms. The transformation in Algorithm~\ref{alg:hs} is indicated by the arrow. Phantom atoms are represented with an asterisk (*) and each atom is associated with an index to identify the atom before and after the transformation: \textbf{a.} atom 1 (\ch{H+}) is marked as phantom; \textbf{b.} atom 0 (\ch{H}) is marked as phantom and it is moved into the attributes of atom 1 (\ch{H}); \textbf{c.} atoms 0 (\ch{^2H}) and 2 (\ch{^2H}) are marked as phantoms and moved into the attributes of atom 1 (\ch{O}); \textbf{d.} atoms 0 (\ch{^2H}) and 2 (\ch{^3H}) are marked as phantoms and moved into the attributes of atom 1 (\ch{N}) which already has one hydrogen attached.}
    \label{fig:step_prepare_hydrogens}
\end{figure}

\begin{figure}[H]
    \centering
    \includesvg[width=\textwidth]{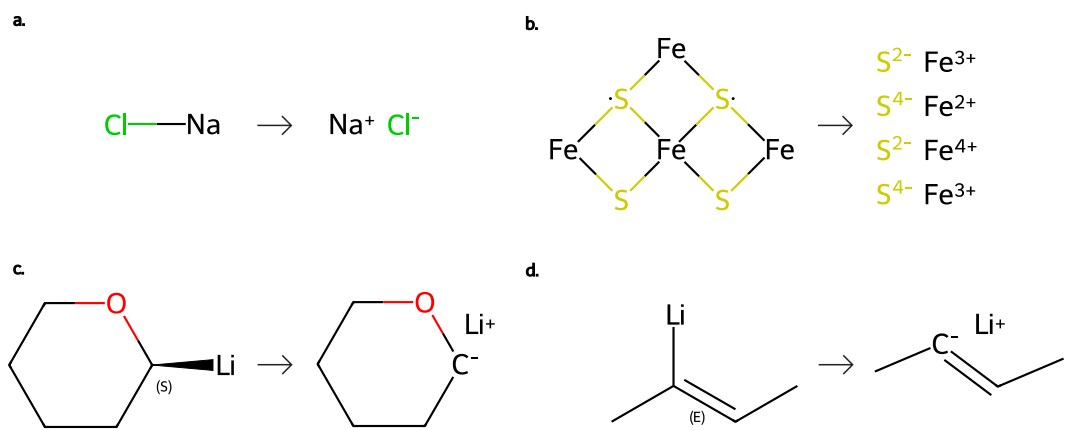}
    \caption{Illustration of Step 2. The transformation in Algorithm~\ref{alg:metals} is indicated by the arrow: \textbf{a.} \ch{NaCl} is disconnected in \ch{Na+} and \ch{Cl-}; \textbf{b.} radicals are replaced by charges; \textbf{c.} stereochemistry is dropped from the disconnected atoms; \textbf{d.} stereochemistry is dropped from the bonds of the disconnected atoms.}
    \label{fig:step_disconnect}
\end{figure}

\begin{figure}[H]
    \centering
    \includesvg[width=\textwidth]{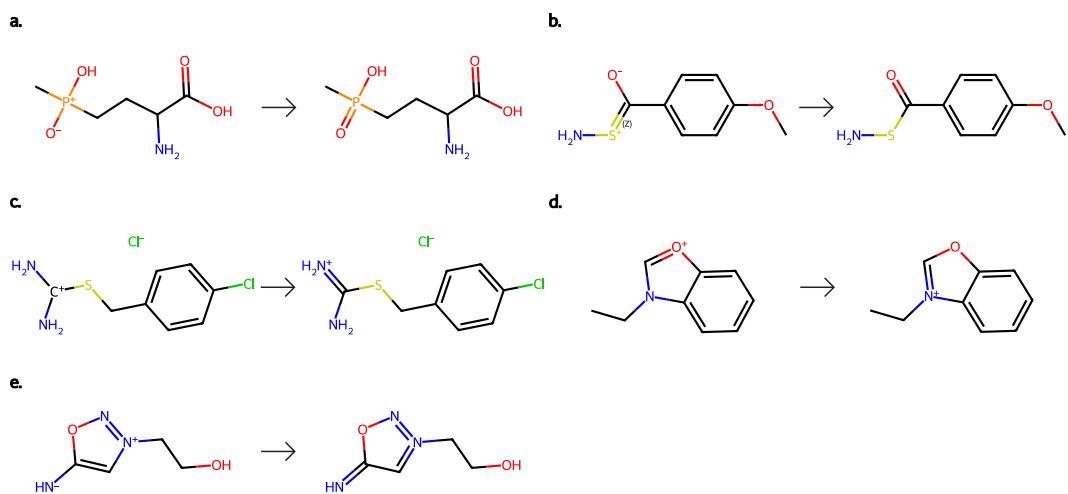}
    \caption{Illustration of Step~3. The  transformations in Table~\ref{tab:step_normalize} are indicated by the arrows: \textbf{a.} transformation~1; \textbf{b.} transformation~2; \textbf{c.} transformation~3; \textbf{d.} transformation~4; \textbf{e.} transformation~5.}
    \label{fig:step_normalize}
\end{figure}

\begin{figure}[H]
    \centering
    \includesvg[width=\textwidth]{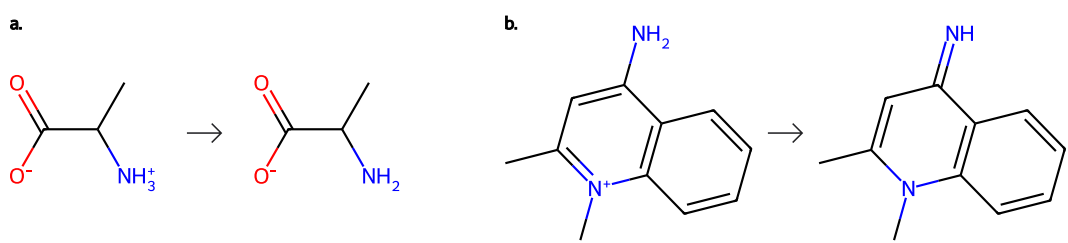}
    \caption{Illustration of Step 4. The transformation is indicated by the arrow: \textbf{a.} removal of \ch{H+} from a charged heteroatom as in~\eqref{eq:soft}; \textbf{b.} removal of \ch{H} from an uncharged heteroatom by changing bonds to \ch{N+} as in~\eqref{eq:hard}.}
    \label{fig:step_deprotonate}
\end{figure}

\begin{figure}[H]
    \centering
    \includesvg[width=\textwidth]{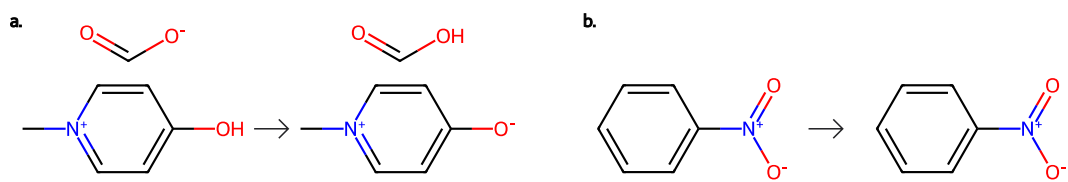}
    \caption{Illustration of Step 5. The transformation is indicated by the arrow: \textbf{a.} each fragment in the molecule is neutralized; \textbf{b.} a neutral fragment remains unaltered.}
    \label{fig:step_neutralize}
\end{figure}

\begin{figure}[H]
    \centering
    \includesvg[width=\textwidth]{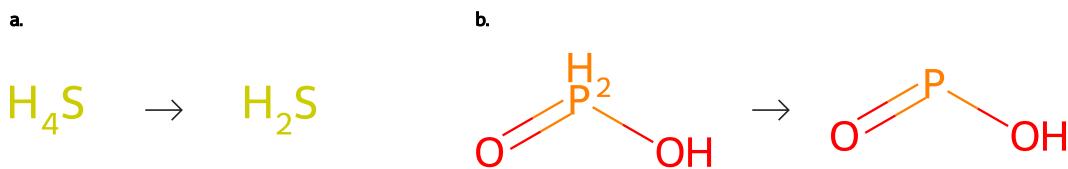}
    \caption{Illustration of Step 6. The transformation is indicated by the arrow: \textbf{a.} valence reduced from 4 to 2; \textbf{b.} valence reduced from 5 to 3.}
    \label{fig:step_drop_h2}
\end{figure}

\begin{figure}[H]
    \centering
    \includesvg[width=\textwidth]{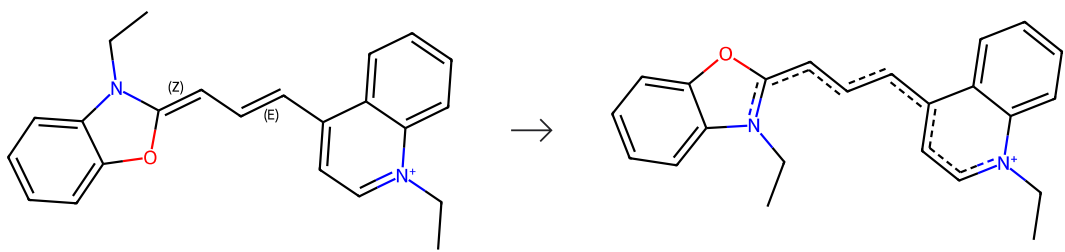}
    \caption{Illustration of Step 7. The bonds in~\eqref{eq:step_mobile_charge} are tagged as ``aromatic" and their stereochemistry is dropped. The transformation is indicated by the arrow.}
    \label{fig:step_mobile_charge}
\end{figure}

\begin{figure}[H]
    \centering
    \includesvg[width=\textwidth]{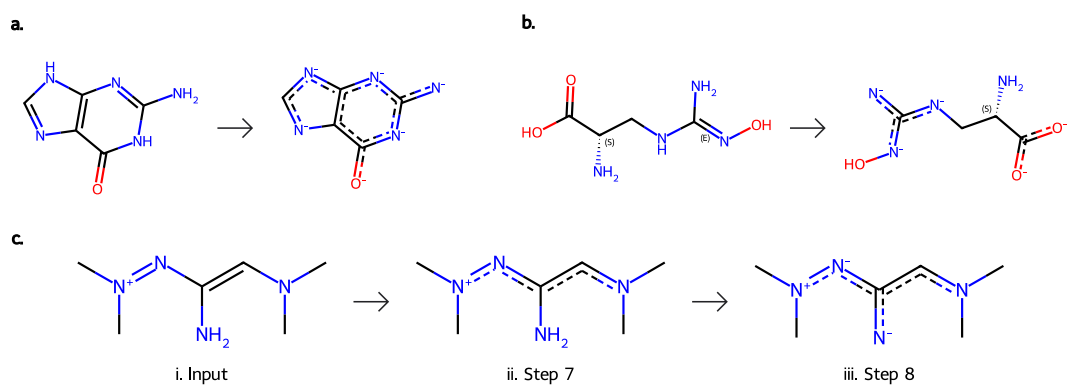}
    \caption{Illustration of Step 8. The transformation in Algorithm~\ref{alg:tautomerism} is indicated by the arrow: \textbf{a.} hydrogens are dropped from tautomeric atoms and replaced with negative charges, tautomeric bonds are tagged as ``aromatic"; \textbf{b.} stereochemistry is dropped from tautomeric bonds; \textbf{c.} the tagging of ``aromatic" bonds in Step~7 enables the discovery of tautomeric patterns in Step~8: i. input graph; ii. graph after Step~7; iii. graph after Step~8.}
    \label{fig:step_mobile_hs}
\end{figure}

\begin{figure}[H]
    \centering
    \includesvg[width=\textwidth]{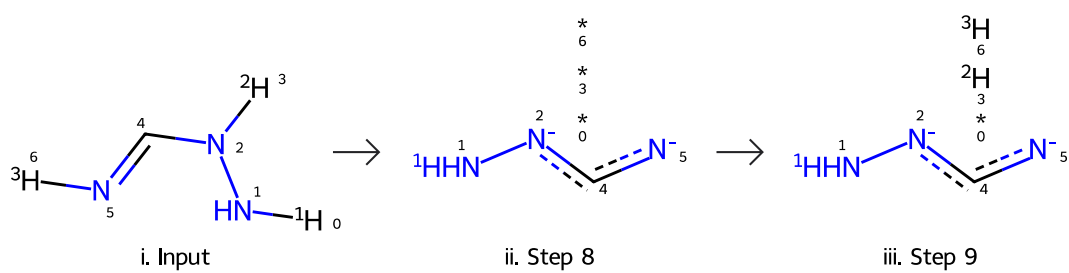}
    \caption{Illustration of Step 9. The phantom tag is removed from mobile hydrogen isotopes: i. input graph, ii. graph after Step 8, iii. graph after Step 9. The transformation is indicated by the arrow.}
    \label{fig:step_isotopes_hs}
\end{figure}

\begin{figure}[H]
    \centering
    \includesvg[width=\textwidth]{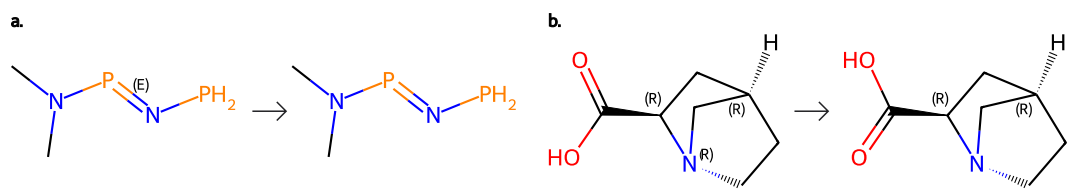}
    \caption{Illustration of Step 10. The transformation is indicated by the arrow. Stereochemistry is dropped in cases not treated as possibly stereogenic by InChI: \textbf{a.} $E/Z$ stereochemistry; \textbf{b.} $R/S$ stereochemistry.}
    \label{fig:step_stereochemistry}
\end{figure}

\begin{figure}[H]
    \includegraphics[width=\linewidth]{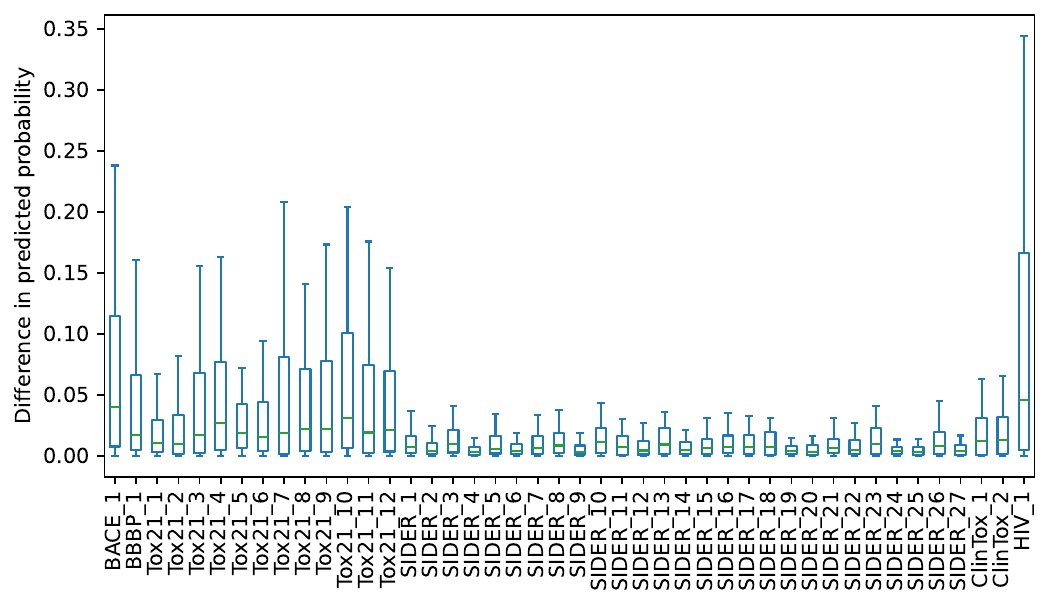}
    \caption{Each boxplot shows the distribution of the difference between the maximum and the minimum predicted probabilities among graphs with the same InChI, for the task specified on the horizontal axis. The box represents the interquartile range, the line inside is the median, and whiskers are the 10th and 90th percentiles. Points beyond the whiskers are not shown.}
    \label{fig:boxplot}
\end{figure}

\section{Supplementary tables}\label{apx:tables}
\renewcommand{\thetable}{C.\arabic{table}}
\setcounter{table}{0}

\begin{table*}[ht]
    \centering
    \caption{Vector of attributes $\mathbf{a}_i$.}
    \label{tab:notation:ai}
    \begin{tabularx}{\textwidth}{l X}
        \toprule
        \textsc{Symbol} & \textsc{Description} \\
        \midrule
        \texttt{Z} & Atomic number. \\
        \texttt{Q} & Formal charge.\\
        \texttt{NumRs} & Number of radical electrons of the atom. \\
        \texttt{NumHs} & Number of hydrogens attached to the atom. \\
        \texttt{Num1H} & Number of protiums attached to the atom. \\
        \texttt{Num2H} & Number of deuteriums attached to the atom. \\
        \texttt{Num3H} & Number of tritiums attached to the atom. \\
        \texttt{SetHs} & Set of hydrogen atoms attached to the atom; default $\{\}$. \\
        \texttt{Degree} &  Number of neighbors of the atom. \\
        \texttt{Isotope} & Isotope number: 0 if natural mixture of isotopes; mass number otherwise.\\
        \texttt{InRing} & Boolean indicating whether the atom is in a ring.\\
        \texttt{Phantom} & Boolean indicating whether the atom must be ignored; default \texttt{False}. \\
        \texttt{ChiralTag} & Integer indicating the RDKit Chiral Type: 0 is none. \\
        \texttt{CIPCode} & Integer indicating tetrahedral stereochemistry: 1 ($R$), -1 ($S$), 0 (none). \\
        \bottomrule
    \end{tabularx}
\end{table*}

\begin{table*}[ht]
    \centering
    \caption{Vector of attributes $\mathbf{b}_{ij}$.}
    \label{tab:notation:bij}
    \begin{tabularx}{\textwidth}{l X}
        \toprule
        \textsc{Symbol} & \textsc{Description} \\
        \midrule
        \texttt{BondOrder} & Bond order: 1 is single; 1.5 is aromatic; 2 is double; 3 is triple. \\
        \texttt{BondStereo} & Integer indicating the RDKit Bond Stereo: 0 is none. \\
        \texttt{BondCIPCode} & Integer indicating double-bond stereochemistry: 1 ($E$), -1 ($Z$), 0 (none). \\
        \bottomrule
    \end{tabularx}
\end{table*}

\begin{table*}
  \caption{Scores and standard errors of neural fingerprints (NeuralFP) and attentive fingerprints (AttentiveFP) on MoleculeNet test sets, using the features in the original papers (Benchmark) and \ii~(InChIfied). Bold values indicate scores that are statistically better between Benchmark and InChIfied models using a two-sided $z$-test at the 95\% confidence level.}
  \label{tab:moleculenet}
  \centering
  \begin{tabular}{lc|cc|cc}
    \toprule
    \textsc{Dataset} & \textsc{Score} & \textsc{NeuralFP} & \textsc{NeuralFP} & \textsc{AttentiveFP} & \textsc{AttentiveFP} \\
     & & \textsc{(Benchmark)} & \textsc{(InChIfied)} & \textsc{(Benchmark)} & \textsc{(InChIfied)} \\
    \midrule
    ESOL & RMSE & 0.70 $\pm$ 0.02  &  0.75 $\pm$ 0.05 & 0.64 $\pm$ 0.04 & 0.59 $\pm$ 0.04 \\
    FreeSolv & RMSE & \textbf{1.13 $\pm$ 0.07}  & 1.47 $\pm$ 0.09 & 1.10 $\pm$ 0.14 & 0.85 $\pm$ 0.13 \\
    Lipo & RMSE & 0.73 $\pm$ 0.02  & 0.73 $\pm$ 0.04 & 0.64 $\pm$ 0.01 & 0.63 $\pm$ 0.01 \\
    HIV & ROC-AUC & 0.80 $\pm$ 0.01  & 0.78 $\pm$ 0.01 & 0.81 $\pm$ 0.01 & 0.79 $\pm$ 0.01 \\
    BACE & ROC-AUC & 0.87 $\pm$ 0.01  &  0.87 $\pm$ 0.01 & 0.88 $\pm$ 0.01 & 0.87 $\pm$ 0.01 \\
    BBBP & ROC-AUC & 0.90 $\pm$ 0.01  &  0.88 $\pm$ 0.02 & 0.90 $\pm$ 0.02 & 0.91 $\pm$ 0.02 \\
    Tox21 & ROC-AUC & 0.82 $\pm$ 0.01  &  0.82 $\pm$ 0.01 &  0.85 $\pm$ 0.00 &  0.84 $\pm$ 0.00 \\
    ToxCast & ROC-AUC & 0.70 $\pm$ 0.01  &  0.71 $\pm$ 0.01 & 0.75 $\pm$ 0.01 & 0.74 $\pm$ 0.01 \\
    SIDER & ROC-AUC & 0.59 $\pm$ 0.01  &  0.60 $\pm$ 0.01 & 0.63 $\pm$ 0.02  & 0.60 $\pm$ 0.01 \\
    ClinTox & ROC-AUC & 0.79 $\pm$ 0.04  &  0.86 $\pm$ 0.02 & 0.91 $\pm$ 0.01 & 0.88 $\pm$ 0.01 \\
    \bottomrule
  \end{tabular}
\end{table*}

\begin{table*}
  \caption{Median runtime by molecular-size bin in milliseconds. MolFromSmiles is the time required to parse the SMILES and construct the molecular graph. ECFP is the time required to generate Extended-Connectivity fingerprints from the graph using standard Daylight invariants. IECFP is the time required to generate the same fingerprints using \ii. The absolute overhead is IECFP $-$ ECFP. The relative overhead is (\textrm{MolFromSmiles} + \textrm{IECFP}) / (\textrm{MolFromSmiles} + \textrm{ECFP}).}
  \label{tab:runtime_by_atoms}
  \centering
  \small
  \setlength{\tabcolsep}{4pt}
  \begin{tabular}{lcccccccccc}
    \toprule
     & \multicolumn{10}{c}{\textsc{Number of Atoms}} \\
    \cmidrule(lr){2-11}
    & \textsc{1--10} & \textsc{11--20} & \textsc{21--30} & \textsc{31--40} & \textsc{41--50} & \textsc{51--60} & \textsc{61--70} & \textsc{71--80} & \textsc{81--90} & \textsc{91--100} \\
    \midrule
    MolFromSmiles & 0.03 & 0.06 & 0.09 & 0.12 & 0.16 & 0.19 & 0.22 & 0.23 & 0.27 & 0.31 \\
    ECFP & 0.02 & 0.03 & 0.04 & 0.06 & 0.07 & 0.09 & 0.10 & 0.13 & 0.14 & 0.16 \\
    IECFP & 0.29 & 0.50 & 0.70 & 0.90 & 1.12 & 1.34 & 1.56 & 1.81 & 2.06 & 2.30 \\
    Abs. Overhead & 0.28 & 0.47 & 0.66 & 0.85 & 1.04 & 1.26 & 1.46 & 1.68 & 1.91 & 2.14 \\
    Rel. Overhead & 6.95 & 6.50 & 5.97 & 5.72 & 5.58 & 5.56 & 5.47 & 5.68 & 5.60 & 5.56 \\
    \bottomrule
  \end{tabular}
\end{table*}

\section{Related Works}\label{sec:background}

\paragraph{International Chemical Identifier.}
Ensuring a unique representation for each molecule is a long-standing challenge in cheminformatics. 
To address this issue, the International Chemical Identifier (InChI)~\cite{heller2015inchi} was developed by IUPAC as a universal standard to encode a molecule into a canonical string: an InChI string is designed so that two chemical structures correspond to the same InChI if and only if they are the same molecule (up to formal chemical rules). InChIs and InChI Keys (hashed representations of InChIs) are widely used to index chemical databases~\cite{bento2020open}, but they have seen limited direct use in machine learning. When used, InChIs typically serve as textual descriptors~\cite{handsel2021translating}, while graph-based models still rely on node and edge features derived from SMILES~\cite{wojtuch2023extended,wang2025recent}. To the best of our knowledge, no prior work has derived node and edge features from the InChI to enforce explanation consistency. 

\paragraph{Chemical invariance in molecular representations.}
Recent work has explored molecular representations that enforce specific chemically motivated invariances. Most closely related, \cite{zalte2025rigr} introduce RIGR, a resonance-invariant graph representation that maps different resonance forms of a molecule to a common representation, thereby reducing sensitivity to an important class of Lewis-structure ambiguity. Other approaches encourage selected invariances, such as tautomeric invariance, through data augmentation by training models on multiple equivalent molecular forms~\cite{ulrich2021exploring}. These methods demonstrate the value of incorporating chemical equivalence into molecular learning, but they typically target a predefined subset of transformations and require either a specialized representation or explicit enumeration of equivalent structures, which can be computationally expensive. In contrast, \ii~use the Standard InChI as the organizing equivalence relation and derive node-, edge-, and graph-level features that incorporate a broader set of chemical standardization operations, including resonance/charge normalization, proton relocation, tautomerism, fragment handling, metal disconnection, and stereochemical conventions. This provides a general featurization framework aligned with an international chemical identity standard.

\paragraph{Model architectures.} 
Modern graph-based models are designed to be invariant to permutation of node indices (graph isomorphisms) \cite{huang2022going}, but they do not solve the broader problem of producing explanations aligned with chemical identity. Similarly, circular fingerprints~\cite{rogers2010extended} and other types of fingerprints~\cite{capecchi2020one} are invariant to atom indexing by hashing local neighborhoods. Neural fingerprints~\cite{duvenaud2015convolutional} replace the hash operation with differentiable operations that can be learnt end-to-end, while ensuring that the output is insensitive to how atoms are ordered. Other works have incorporated equivariance to physical transformations. For example, \cite{schutt2017schnet} introduces a network that operates on 3D atomic coordinates and guarantees rotationally invariant predictions for molecular energy. Yet, no existing model architecture can enforce invariance with respect to all transformations of the molecular graph that do not alter its chemical identity.

\paragraph{Explanation methods.}
General explainability methods such as LIME~\cite{ribeiro2016should} and SHAP~\cite{lundberg2017unified} have become important tools to interpret model predictions. Other techniques specifically designed for deep learning include Integrated Gradients~\cite{sundararajan2017axiomatic}, DeepLIFT~\cite{shrikumar2017learning}, saliency maps~\cite{simonyan2013deep}, guided backpropagation~\cite{springenberg2014striving}, and GradCAM~\cite{selvaraju2017grad}, among others. For graph neural networks, GNNExplainer~\cite{ying2019gnnexplainer} and PGExplainer~\cite{luo2020parameterized} identify important subgraphs contributing to predictions. Another approach is to incorporate interpretability by design. For instance, AttentiveFP~\cite{xiong2019pushing} incorporates attention weights that provide a form of explanation by indicating which parts of the molecule contribute most to the prediction. Despite the diversity of approaches, a common limitation is that different inputs representing the same molecule generally lead to different explanations. 

\section{Construction of the alternative-depiction datasets}\label{sec:augmentation}

Conceptually, the ideal experiment would enumerate many valid SMILES strings for each InChI. However, because there is no general tool for exhaustive enumeration of all SMILES representations associated with a given InChI, we use the round-trip conversion SMILES$\rightarrow$InChI$\rightarrow$SMILES as a practical procedure for generating an alternative depiction. For each SMILES string in the original MoleculeNet datasets, we compute its InChI and convert the InChI back to SMILES. If the round-trip SMILES differs from the original SMILES, we add it to the corresponding alternative-depiction dataset and assign it the same label as the original SMILES. This procedure yields molecular graphs that are not present in the original datasets, but that preserve the same InChI and therefore the same chemical identity. Cases where the round-trip conversion returns the original SMILES are discarded, since they do not provide an alternative graph depiction.

\section{Phantom atoms in neural networks}\label{sec:phantom}

Phantom atoms are retained to preserve the original atom indexing, which simplifies mapping attributions back to the input graph. In our experiments, they are represented with zero-valued features. Under the Integrated Gradients setup used in this work, zero-valued phantom atoms receive zero attribution; however, zero attribution does not necessarily imply zero effect on the prediction. Therefore, for applications in which phantom atoms are present, they should be generally removed from the neural-network input by reindexing atoms and edges, and used only as a bookkeeping device for visualization. In our MoleculeNet experiments, phantom atoms occur in only 0.07\% of molecular graphs, so this implementation detail affects a negligible fraction of samples.

\section{Long-range tautomerisms}\label{apx:keto}

From Table~\ref{tab:confusion}, we can also observe that, when the InChI Keys are different, \ii~produced identical fingerprints 40 times more often (13,199 vs. 335). The reason why these particular pairs collide more often is that InChI, by design, does not capture all possible tautomeric forms. As stated in the InChI Technical FAQ 6.4 (\url{https://www.inchi-trust.org/technical-faq/}): ``In its current state, InChI recognizes the most common form of H migration (for the full list, see Table 6, Section IVb of the InChI Technical Manual). However, several ways of tautomeric migration that are not supported by default may appear important for some chemists. In particular, these are keto-enol and long-range tautomerisms.'' Instead, our method can handle some forms of long-range tautomerisms, which means that some graph pairs considered distinct by InChI are treated as equivalent by our~\ii. We manually inspected 100 random collisions and found that 94 of them were due to the different handling of tautomerisms.

\section{Licenses of existing assets}\label{apx:license}

The results in this paper were generated using the following open-source software and open data: RDKit (BSD license)~\cite{rdkit}, PyTorch (BSD license)~\cite{pytorch}, Pandas (BSD license)~\cite{pandas}, NumPy (BSD license)~\cite{numpy}, Captum (BSD license)~\cite{captum}, Matplotlib (Python Software Foundation License)~\cite{matplotlib}, MoleculeNet~\cite{wu2018moleculenet} (MIT License), and PubChem Substances~\cite{kim2016pubchem,kim2025pubchem}. The Fair Use Disclaimer of PubChem Substances is available at \url{https://ftp.ncbi.nlm.nih.gov/pubchem/Substance/README} and is reported hereafter for completeness: ``Databases of molecular data on the NCBI FTP site include examples as nucleotide sequences (GenBank), protein sequences, macromolecular structures, molecular variation, gene expression, and mapping data. They are designed to provide and encourage access within the scientific community to sources of current and comprehensive information. Therefore, NCBI itself places no restrictions on the use or distribution of the data contained therein. However, some submitters of the original data may claim patent, copyright, or other intellectual property rights in all or a portion of the data they have submitted. NCBI is not in a position to assess the validity of such claims and, therefore, cannot provide comment or unrestricted permission concerning the use, copying, or distribution of the information contained in the molecular databases.''


\end{document}